\documentclass[runningheads]{llncs}

\usepackage[mobile]{eccv}
\usepackage{multirow}
\usepackage{marvosym}
\usepackage[accsupp]{axessibility}
\makeatletter

\usepackage{eccvabbrv}

\usepackage{graphicx}
\usepackage{booktabs}
\usepackage{algorithm}
\usepackage{algorithmic}
\usepackage[accsupp]{axessibility}  
\usepackage{amsmath,amsfonts,bm}

\usepackage[pagebackref,breaklinks,colorlinks]{hyperref}
\usepackage{orcidlink}

\begin{document}

\title{DreamReward: Text-to-3D Generation with Human Preference} 


\author{JunLiang Ye\inst{1,2}\textsuperscript{\dag} \and
Fangfu Liu\inst{1}\textsuperscript{\dag} \and
Qixiu Li\inst{1} \and
Zhengyi Wang\inst{1,2} \and
Yikai Wang\inst{1} \and
Xinzhou Wang\inst{1,2} \and Yueqi Duan\inst{1}\textsuperscript{\Letter} \and  Jun Zhu\inst{1,2}\textsuperscript{\Letter} }
\authorrunning{Ye et al.}



\newcommand\blfootnote[1]{%
\begingroup
\renewcommand\thefootnote{}\footnote{#1}%
\addtocounter{footnote}{-1}%
\endgroup
}
\institute{Tsinghua University \and ShengShu, Beijing, China \blfootnote{\dag \ Equal contribution.}\blfootnote{\Letter \ Corresponding authors.}}
\maketitle
\vspace{-5mm}
\begin{abstract}
3D content creation from text prompts has shown remarkable success recently. However, current text-to-3D methods often generate 3D results that do not align well with human preferences. In this paper, we present a comprehensive framework, coined \textbf{DreamReward}, to learn and improve text-to-3D models from human preference feedback. To begin with, we collect 25k expert comparisons based on a systematic annotation pipeline including rating and ranking. Then, we build Reward3D---the first general-purpose text-to-3D human preference reward model to effectively encode human preferences. Building upon the 3D reward model, we finally perform theoretical analysis and present the Reward3D Feedback Learning (DreamFL), a direct tuning algorithm to optimize the multi-view diffusion models with a redefined scorer. 
Grounded by theoretical proof and extensive experiment comparisons, our DreamReward successfully generates high-fidelity and 3D consistent results with significant boosts in prompt alignment with human intention. Our results demonstrate the great potential for learning from human feedback to improve text-to-3D models. Project Page: \url{https://jamesyjl.github.io/DreamReward/}.
\vspace{-2mm}
\end{abstract}
\section{Introduction}
\label{sec:intro}
3D content generation has wide applications in various fields (\eg, films, animation, game design, architectural design, and virtual reality). In recent years, significant advancements in diffusion models have greatly propelled the development of automated 3D generation. 3D creation can be classified into two principal categories\cite{tang2023dreamgaussian}: inference-only 3D native methods\cite{jun2023shape,gupta20233dgen} and optimization-based 2D lifting methods \cite{poole2022dreamfusion,wang2023prolificdreamer,chen2023fantasia3d,lin2023magic3d,kerbl20233d,chen2023gaussianeditor,tang2024lgm,hong2023lrm}. Given a text or an image, these models are capable of generating highly intricate 3D content, some even overcoming multi-face issues to produce high-quality and viewpoint-consistent 3D models\cite{shi2023mvdream,liu2023zero1to3,liu2023one2345}. 
Despite rapid advancements, some researchers\cite{shi2023mvdream}indicate that the 3D content generated by existing generative models struggles to align with human preferences. Typically, this inconsistency includes but is not limited to text-3D alignment, overall quality, and multi-view consistency.

Recently, some works have applied reinforcement learning from human feedback (RLHF) to natural language processing (NLP)\cite{stiennon2022learning,ouyang2022training} and text-to-image generation\cite{xu2023imagereward,wada2024polos,zhu2024diffusion,black2024training}. These algorithms typically begin by constructing and annotating datasets based on human feedback, and then training reward models. Finally, they finetune large models (such as large language models or diffusion models) using reinforcement learning techniques. This allows the fine-tuning models to better align with human preferences. 

Inspired by the aforementioned works, we recognize the effectiveness of RLHF in improving the performance of generative models. In this work, we propose \textbf{DreamReward}, which greatly boosts high-text alignment and high-quality text-to-3D generation through human preference feedback. We propose the first general-purpose human preference reward model for text-to-3D generation, Reward3D. First, we use a clustering algorithm to extract 5k of the most representative prompts from Cap3D\cite{luo2023scalable} and generate a corresponding 3D dataset. Subsequently, we produce 10 3D contents for each prompt and filtered them based on quality, resulting in 2530 prompt sets, where each prompt corresponds to 4$\sim$10 3D contents. After collecting 25k pairs of expert comparisons, we annotate the comparison and trained the Reward3D model based on it.

 After constructing the annotated 3D Dataset, we train a 3D-aware scoring model for text-to-3D generation on the constructed 3D dataset. Given the most representative 110 prompts generated by GPTEval3D \cite{wu2024gpt4vision} and compared to the 2D scoring models ImageReward \cite{xu2023imagereward} and CLIP \cite{radford2021learning}, which lack 3D-aware capabilities, our Reward3D can consistently align with human preference ranking and exhibit higher distinguishability among different 3D models. With its high alignment in human preference observed from experiments, we suggest that Reward3D could serve as a promising automatic text-to-3D evaluation metric.
 
 Building upon this, we further explore an optimization approach to improve 3D generation results—Reward3D Feedback Learning (DreamFL), which is a direct tuning algorithm designed to optimize multi-view diffusion models using a redefined scorer. Based on Reward3D, we carefully design the LossReward and incorporate it into the SDS pipeline for 3D generation. Grounded by our mathematical derivation, we find that the LossReward effectively drives the optimization of 3D models towards higher quality and alignment. Extensive experimental results demonstrate that 3D assets generated by DreamFL not only achieve impressive visualization but also outperform other text-to-3D generation methods in terms of quantitative metrics such as GPTEval3D \cite{wu2024gpt4vision}, CLIP \cite{radford2021learning}, ImageReward \cite{xu2023imagereward}, and our Reward3D.

To summarise, we make the following contributions:
\begin{itemize}    
    \item \textbf{Labeled-3D dataset}: We are among the first to construct and annotate a diverse 3D dataset suitable for training and testing models aligned with human preferences.
    \item \textbf{Reward3D}: We train the Reward3D scoring model with 3D-aware capabilities, enabling it to effectively evaluate the quality of generated 3D content. 
    \item \textbf{DreamFL}: We propose the Reward3D Feedback Learning (DreamFL) algorithm to enhance the human preference alignment in our 3D results. 

\end{itemize}

\section{Related Work}
\label{sec:relat}
\subsection{Text-to-image Generation}
\label{sec:relat_1}
Diffusion models~\cite{sohldickstein2015deep,ho2020denoising,dhariwal2021diffusion} combining with large-scale language encoders~\cite{radford2021learning,48643}, have become the leading approach in text-to-image generation. Typically, they involve a two-step process. Initially, noise is progressively added to the original data until it aligns with a prior distribution, such as the Gaussian distribution. Subsequently, a neural network is employed to predict the previously added noise, allowing the initialized samples from the prior distribution to undergo a step-by-step reverse denoising process. Leveraging this technique, recent advancements~\cite{rombach2022highresolution, nichol2022glide, ramesh2022hierarchical} have demonstrated the ability to synthesize images of exceptional quality.
\subsection{Text-to-3D Generation}
\label{sec:relat_2}
With the high development of text-to-image diffusion models, there has been a surge of studies in text-to-3D generation recently. Due to limited diverse 3D datasets \cite{chang2015shapenet} compared to 2D, DreamFusion\cite{poole2022dreamfusion} and SJC\cite{wang2022score} have shifted towards exploring the route of distilling score from 2D diffusion priors to optimizes a 3D representation such as NeRF\cite{mildenhall2020nerf}, and show very promising results. Such distillation-based methods\cite{poole2022dreamfusion,wang2023prolificdreamer,shi2023mvdream,chen2023fantasia3d,lin2023magic3d,wang2023luciddreaming,wei2023adversarial,zhu2023hifa,zhuang2023dreameditor} have undergone rapid improvements in recent years. However, there still exists a significant gap in the generation quality between 3D and 2D generation. 3D generation still lacks in terms of generation speed, diversity of themes, and alignment with human preferences. To improve the efficiency of 3D generation, DreamGaussian\cite{chen2023gaussianeditor} transformed the 3D representation from NeRF to gaussian-splatting\cite{kerbl20233d}, resulting in high-quality generation effects. On the other hand, Shap-E\cite{jun2023shape} and Point-E\cite{nichol2022pointe} achieve 3D generation in a matter of minutes through pretraining on massive undisclosed 3D datasets.
In this work, we will continue to narrow the gap between 3D and 2D generation through the DreamReward framework, which guides 3D models towards high-quality and highly aligned generation.
\subsection{Text-to-3D Generation Evaluation Metrics.}
\label{sec:relat_3}
Evaluating text-to-3D generation models is a highly challenging task, requiring both 3D awareness and understanding of textual semantics.
The existing text-to-3D evaluation methods mainly include approaches that utilize multimodal embeddings, such as CLIP\cite{jain2022zeroshot,radford2021learning} and BLIP\cite{li2022blip,li2023blip2}, as well as methods, such as GPTEval3D\cite{wu2024gpt4vision} and T3batch\cite{he2023t3bench} that employ large-scale multimodal language models GPT-4V~\cite{achiam2023gpt-4v}. To obtain a fair and reliable collection of text prompts, GPTEval3D \cite{wu2024gpt4vision} created a text-prompt generator using language instruction. This enables us to better assess the strengths and weaknesses of different 3D models. Our proposed Reward3D in this work will serve as a novel evaluation model for text-to-3D, assisting users in evaluating 3D results effectively without relying on large language models, which may incur costs. It will also provide promising scores and rankings aligned with human preferences.
\subsection{Learning from human feedback}
\label{sec:relat_4}
The alignment of large language models (LLMs)\cite{openai2023gpt4, geminiteam2023gemini} with human preferences is an issue that has garnered considerable attention. Reinforcement Learning from Human Feedback (RLHF)\cite{ouyang2022training, stiennon2022learning, ziegler2020finetuning} uses a strategy that leverages human feedback with reinforcement learning policies to address this challenge. Recent literature\cite{xu2023imagereward, yang2023using, black2024training, wallace2023diffusion, fan2023dpok, yang2023using} has demonstrated that incorporating human feedback enhances the performance of text-to-image models as well. ImageReward\cite{xu2023imagereward} introduces a reward model based on human preferences specifically for text-to-image tasks and puts forward a novel approach termed Reward Feedback Learning, which is designed to refine diffusion models. Meanwhile, DiffusionDPO\cite{wallace2023diffusion} presents a technique that aligns diffusion models with human preferences by directly optimizing with human comparative data, an adaptation of Direct Preference Optimization (DPO)\cite{rafailov2023direct}. Further, DPOK\cite{fan2023dpok} amalgamates policy optimization with KL regularization within the framework of text-to-image diffusion models. Despite the proven efficacy of these approaches, learning from human feedback in the domain of text-to-3D generation still requires investigation. 
\section{Overall Framework}
\label{sec:Overall Framework}
\begin{center}
\begin{figure}[tb]
  \centering
  \includegraphics[width=1\linewidth]{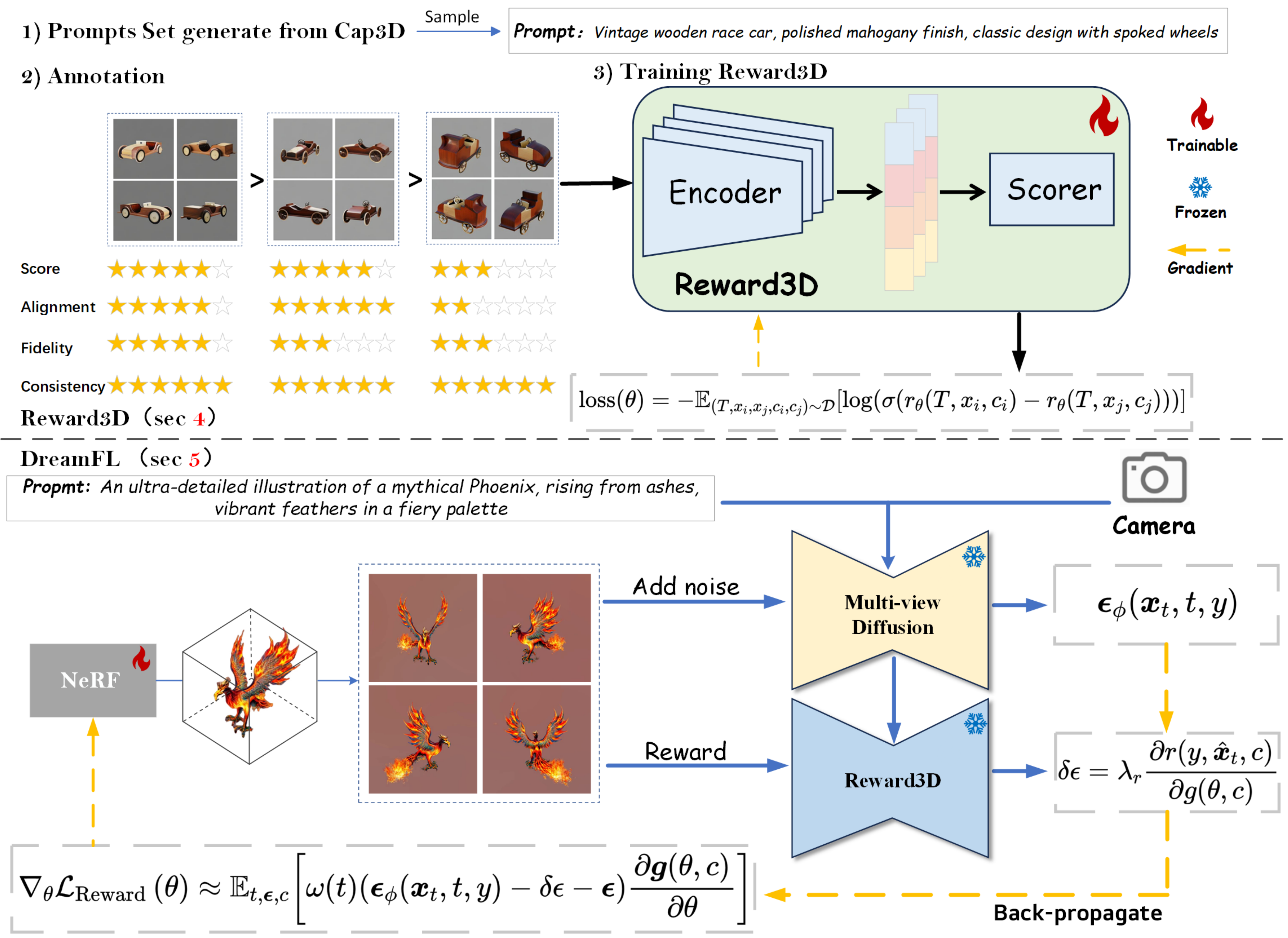}
  \caption{\textbf{The overall framework of our DreamReward}. (\textit{Top}) Reward3D involves data collection, annotation, and preference learning. (\textit{Bottom}) DreamFL utilizes feedback from Reward3D to compute RewardLoss and incorporate it into the SDS loss for simultaneous optimization of NeRF.}
  \label{fig:pipeline}
\end{figure}
\end{center}
\vspace{-0.5em}
We hereby present the overall framework of our DreamReward, a novel text-to-3D framework to achieve human preference alignment. The complete pipeline is depicted in Figure~\ref{fig:pipeline}. Initially, in Sec.~\ref{sec:Reward3D}, we introduce the Reward3D architecture, which encompasses the construction of a 3D dataset (Sec.~\ref{sec:Reward3D_1}), the development of a data annotation pipeline (Sec.~\ref{sec:Reward3D_2}), and the training of the Reward Model (RM) (Sec.~\ref{sec:Reward3D_4}). After training the 3D-aware Reward3D model, we proceed to delineate the core of our DreamReward framework—the Reward3D Feedback Learning (DreamFL) in Sec.~\ref{sec:DreamFL}.  Specifically, In Sec.~\ref{sec:5.2}, we first identify that the distribution obtained by existing diffusion models for distilling 3D assets diverges from the desired distribution at two distinct levels. Then in Sec.~\ref{sec:5.3}, we demonstrate the efficacy of our Reward3D in bridging the gap between these two distributions through both detailed mathematical derivation and demonstration. More detailed implementation specifics of our algorithm can be found in our supplementary materials.

\section{Reward3D}
\label{sec:Reward3D}
\subsection{Annotation Pipeline Design}
\subsubsection{Prompt Selection and 3D Collection.}
\label{sec:Reward3D_1}
Our proposed new dataset utilizes a diverse selection of prompts from cap3D\cite{luo2023scalable}, which is a re-annotation of the large-scale 3D dataset Objaverse \cite{deitke2022objaverse}, with better alignment compared to the original prompts in Objaverse \cite{deitke2022objaverse}. To ensure diversity in selected prompts, we employ a graph-based algorithm that leverages language model-based prompt similarity. This selection yields 5000 candidate prompts, each accompanied by 4-10 sampled 3D assets generated from \emph{ashawkey/mvdream-sd2.1-diffusers} \cite{shi2023mvdream}
\subsubsection{Dataset Filtering.}
\label{sec:Reward3D_2}
From our empirical observations, we notice that results we generate are prone to encountering mode collapses. This is attributed to the complexity of the selected prompts~\cite{wu2024gpt4vision}, leading to significant collapses of the corresponding 3D assets under the same prompt. Therefore, prior to annotation, we conduct a filtering process on the generated 3D dataset with a selection of 2530 prompts, each from 4-10 assets. Then we obtain 25,304 candidate pairs for labeling.
\subsubsection{Human Annotation Design.}
\label{sec:Reward3D_3}
In the annotation process, annotators rate images on a scale of 1-6 based on text-3D alignment, overall quality, and multi-view consistency, and rank them according to the average scores. To avoid potential conflicts in rankings, we maintain a real-time data structure for each prompt set. When conflicts arise, conflicting pairs are flagged, followed by secondary verification and correction. We recruit some annotators from universities for labeling, and additionally sought assistance from data institutions, with annotation documents showcased in the appendix.
\subsection{Reward3D Training}
\label{sec:Reward3D_4}
Similar to RM training for language model of previous works\cite{ouyang2022training, stiennon2022learning, ziegler2020finetuning}, we formulate the preference annotations as rankings. We have 9 3D-model ranked for the same prompt T, and get at most $C_9^2$ comparison pairs if there are no ties between two 3D-model. For each comparison, if $x_i$ is better and $x_j$ is worse, the loss function can be formulated as:
\begin{equation}
\label{eq1}
\operatorname{loss}(\theta)=-\mathbb{E}_{\left(T, x_i, x_j, c_i, c_j\right) \sim \mathcal{D}}\left[\log \left(\sigma\left(r_\theta\left(T, x_i, c_i\right)-r_\theta\left(T, x_j, c_j\right)\right)\right)\right],
\end{equation}
where $r$ represents the Reward3D model, $c_i$ and $c_j$ represent cameras.
\subsubsection{Training Detail.}We use ImageReward \cite{xu2023imagereward} as the backbone of our Reward3D. 
We extract image and text features, combine them with cross-attention, and use an MLP to generate a scalar for preference comparison. During the training stage, we observe rapid convergence and consequent overfitting, which harms its performance. To address this, we freeze some backbone transformer layers’ parameters, finding that a proper number of fixed layers can improve the model’s performance. We train Reward3D on a single 4090 GPU (24GB) with a batch size set to 8. We utilize the AdamW\cite{loshchilov2019decoupled} optimizer with a learning rate of 1e-5 and a fixed rate set to 80\%.

\section{DreamFL}
\label{sec:DreamFL}
After training the 3D reward model, we now present DreamFL algorithm. Our pipeline is depicted in Figure~\ref{fig:pipeline}. Before delving into the specifics of our approach, let's start by revisiting the preliminaries of Score Distillation Sampling theory.
\subsection{Preliminaries}
\subsubsection{Score Distillation Sampling (SDS).}
Score Distillation Sampling (SDS)\cite{poole2022dreamfusion}, an optimization method that distills 3D knowledge from pretrained 2D diffusion models, has significantly advanced the rapid development of 3D generation \cite{poole2022dreamfusion,wang2023prolificdreamer,lin2023magic3d,wang2023luciddreaming,zhu2023hifa} in recent years. Given a differentiable rendering mapping function g($\theta$,c), a pretrained 2D diffusion model $\phi(x_t|y)$ and its corresponding noise prediction network $\epsilon_{\phi}(x_t,t,y)$, SDS optimizes the parameter $\theta$ by solving:
\begin{equation}
\label{eq2}
\nabla_\theta \mathcal{L}_{\mathrm{SDS}}(\theta) \approx \mathbb{E}_{t, \boldsymbol{\epsilon}, c}\left[\omega(t)\left(\boldsymbol{\epsilon}_{\phi}\left(\boldsymbol{x}_t, t, y\right)-\boldsymbol{\epsilon}\right) \frac{\partial \boldsymbol{g}(\theta, c)}{\partial \theta}\right],
\end{equation}
where $\theta$ is the parameter of 3D representation and c is desired camera. To elaborate further, we denote $q_t^\theta\left(\boldsymbol{x}_t|c\right)$ as the distribution at time $t$ of the forward diffusion process, initiated from the rendered image. The SDS optimization algorithm described above can be regarded as
\begin{equation}
\label{eq3}
\min _{\theta \in \Theta} \mathcal{L}_{\mathrm{SDS}}(\theta):=\mathbb{E}_{t, c}\left[\left(\sigma_t / \alpha_t\right) \omega(t) D_{\mathrm{KL}}\left(q_t^\theta\left(\boldsymbol{x}_t|c\right) \| p_t\left(\boldsymbol{x}_t|y\right)\right)\right].
\end{equation}

\subsection{DreamFL}
\label{sec:5.2}
\subsubsection{Why do SDS-generated 3D assets lack alignment?}
Two major challenges arise when seeking to align 3D generation models with human preferences. (1) the conditional distribution $p_t\left(\boldsymbol{x}_t|y\right)$ obtained by pre-trained diffusion models diverges from human preference and actual user-prompt distributions\cite{yang2023using, xu2023imagereward}. Consequently, 3D assets distilled from this deviant distribution inherently fail to align with human preferences, often to an even more pronounced degree.
(2)The capability to maintain multi-view consistency acquired through score distillation from diffusion models is profoundly limited\cite{ding2023textto3d,li2023sweetdreamer,liu2023sherpa3d}.

In recent years, many related works\cite{black2024training, wallace2023diffusion} have emerged in the field of text-to-image generation to address the aforementioned problem (1). However, due to the multi-step nature of the process, fine-tuning a diffusion model suitable for the text-to-3D domain is challenging. The main difference lies in the timestamp, where fine-tuning a diffusion model for generation typically requires fine-tuning the last 20\%-30\% of denoising steps\cite{xu2023imagereward}. In contrast, text-to-image tasks often only require the last 10 steps for fine-tuning, as they use around 40 steps. Through empirical studies, it is found that 3D generation often requires over 1000 denoising steps\cite{threestudio2023,wang2023prolificdreamer}, meaning over 20 times the computational workload. On the other hand, the inherent multi-view inconsistency of the SDS algorithm will exacerbate the perceived quality of the generated 3D results among humans. Suppose there exists a diffusion model $\hat{\phi}$ that aligns well with human preferences. However, the presence of problem (2). causes it to lack awareness of 3D, resulting in the generated 3D assets still not meeting our desired criteria. Based on the above analysis, we have learned that the distribution $p_t\left(\boldsymbol{x}_t|y\right)$, followed by our current diffusion model, deviates significantly from the ideal distribution $p^r_t\left(\boldsymbol{x}_t|y\right)$, where $p^{r}_t\left(\boldsymbol{x}_t|y\right)$ represents the target distribution aligned with human preferences and possessing 3D awareness.

To address these challenges, our approach aims to leverage our existing distribution $p_t\left(\boldsymbol{x}_t|y\right)$ to approximate the challenging distribution $p^r_t\left(\boldsymbol{x}_t|y\right)$. Inspired by ProlificDreamer, which used a LoRA \cite{zhang2023adding} to approximate the distribution of NeRF, we found that approximating the predicted noise of a distribution is sufficient to approximate the distribution itself. Therefore, we denote the noise generated from distributions $p_t\left(\boldsymbol{x}_t|y\right)$ and $p^r_t\left(\boldsymbol{x}_t|y\right)$ as $\epsilon_{\phi}(x_t,t,y)$ and $\epsilon^{r}_{\phi}(x_t,t,y)$, respectively. Our goal is to approximate $\epsilon^{r}_{\phi}(x_t,t,y)$ using $\epsilon_{\phi}(x_t,t,y)$, meaning to obtain $\delta \epsilon = \epsilon_{\phi}(x_t,t,y)-\epsilon^{r}_{\phi}(x_t,t,y)$, which will effectively address the problem. With this understanding, we proceed to the derivation below.
\subsubsection{Approximate $\delta \epsilon$ using Reward3D.}
First, let us rewrite Eq.~\ref{eq3}
by replacing $p_t\left(\boldsymbol{x}_t|y\right)$ with $p^{r}_t\left(\boldsymbol{x}_t|y\right)$. 
\begin{equation}
\label{eq4}
\min _{\theta \in \Theta} \mathcal{L}_{\mathrm{Reward}}(\theta):=\mathbb{E}_{t, c}\left[\left(\sigma_t / \alpha_t\right) \omega(t) D_{\mathrm{KL}}\left(q_t^\theta\left(\boldsymbol{x}_t|c\right) \| p^{r}_t\left(\boldsymbol{x}_t|y\right)\right)\right].
\end{equation}
Its gradient is approximated by
\begin{equation}
\label{eq5}
\nabla_\theta \mathcal{L}_{\mathrm{Reward}}(\theta) \approx \mathbb{E}_{t, \boldsymbol{\epsilon}, c}\left[\omega(t)\left(\boldsymbol{\epsilon}^r_{\phi}\left(\boldsymbol{x}_t, t, y\right)-\boldsymbol{\epsilon}\right) \frac{\partial \boldsymbol{g}(\theta, c)}{\partial \theta}\right].
\end{equation}
However, as mentioned above, obtaining the distribution $p^{r}_t\left(\boldsymbol{x}_t|y\right)$ and  $\epsilon^{r}_{\phi}(x_t,t,y)$ is a very challenging task. To overcome this challenge, we approximate $\epsilon^{r}_{\phi}(x_t,t,y)=\epsilon_{\phi}(x_t,t,y)-\delta \epsilon$ using the pretrained $\epsilon_{\phi}(x_t,t,y)$ prediction network and Reward3D.
Let $\boldsymbol{g}(\theta, c) = x\in \mathbb{R}^{4\times H \times W \times C}$ denote the multiview images with continuous angles spanning 360 degrees, $\boldsymbol{x}_t=\alpha_t \boldsymbol{x}+\sigma_t \boldsymbol{\epsilon}$ and $\hat{\boldsymbol{x}}_t=\frac{1}{\alpha_t}\left[\boldsymbol{x}_t-\sigma_t \epsilon_{\phi}(x_t,t,y)\right] $ denote the prediction. The difference $\delta \epsilon$ between the $\epsilon^r_{\phi}(x_t,t,y)$  and $\epsilon_{\phi}(x_t,t,y)$  can be calculated by:
\begin{equation}
\label{eq6}
\delta \epsilon=-\frac{\partial{r(y,\hat{\boldsymbol{x}}_t,c)}}{\partial{\epsilon_{\phi}(x_t,t,y)}}=\lambda_r \frac{\partial{r(y,\hat{\boldsymbol{x}}_t,c)}}{\partial{g(\theta,c)}},
\end{equation}
Where r stands for Reward3D, and c stands for camera. Therefore, Eq.~\ref{eq5} can be reorganized as:
\begin{equation}
\label{eq7}
\nabla_\theta \mathcal{L}_{\mathrm{Reward}}(\theta) \approx \mathbb{E}_{t, \boldsymbol{\epsilon}, c}[\omega(t)( \underbrace{\boldsymbol{\epsilon}_{\phi}\left(\boldsymbol{x}_t, t, y\right)-\lambda_r \frac{\partial{r(y,\hat{\boldsymbol{x}}_t,c)}}{\partial{g(\theta,c)}}}_{\boldsymbol{\epsilon}_{\phi}^r\left(\boldsymbol{x}_t, t, y\right)}-\boldsymbol{\epsilon}) \frac{\partial \boldsymbol{g}(\theta, c)}{\partial \theta}].
\end{equation}
Eq.~\ref{eq4} can be reorganized as:
\begin{equation}
\label{eq8}
\mathcal{L}_{\mathrm{Reward}}(\theta) \approx \mathcal{L}_{\mathrm{SDS}}(\theta)-\lambda_{r} r(y,\hat{\boldsymbol{x}_t},c).
\end{equation}
\subsection{Implementation Details}
\label{sec:5.3}
We use MVDream\cite{shi2023mvdream} as our backbone, which is capable of generating multi-view consistent 3D assets. MVDream can
align well with our multi-view optimization pipieline. For weighting factors, we define $\lambda_r$ as t($\mathcal{L}_{\mathrm{SDS}}$, $r(y,\hat{\boldsymbol{x}_t},c)$) multiplied by $\mu$, where t is a weighting function used to ensure consistency in the magnitudes of $\mathcal{L}_{\mathrm{SDS}}$ and $r(y,\hat{\boldsymbol{x}_t},c)$. $\mu$ increases from 0 to 0.25 throughout the training process until reaching 0.6. Upon completion of training, we deactivate the $\mathcal{L}_{\mathrm{SDS}}$ and solely fine-tune the 3D results for 200 steps only using $r(y,\hat{\boldsymbol{x}_t},c)$ with a large $\mu$. Experimental results demonstrate that this approach enhances the aesthetics and stability of the training process. Please refer to our supplementary materials for more implementation details.
\begin{figure}[tb]
  \centering
  \includegraphics[width=1\linewidth]{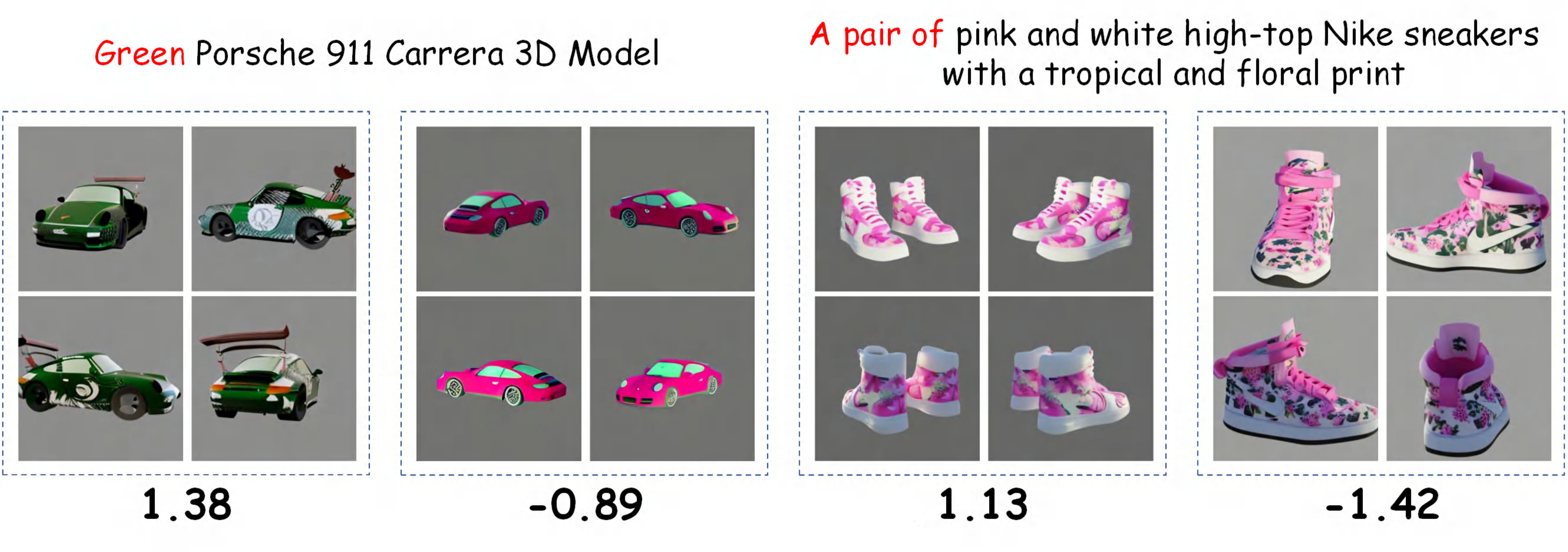}
  \caption{Representative examples from our constructed 3D dataset, along with the scores assigned by Reward3D. Reward3D gives lower scores to 3D assets deviating from the prompt description.}
  \label{fig:a1}
\end{figure}

\section{Experiments}
\label{sec:Experiments}
In this section, we conduct extensive experiments to evaluate our text-to-3D method DreamFL and text-to-3D evaluation model Reward3D. In Sec.~\ref{sec:6.2}, We first present qualitative results compared with five baselines. Then we report the quantitative results with four evaluation metrics and user studies. All of this indicates that our DreamReward model beats the other five models and conforms to human preference. In Sec.~\ref{sec:6.3}, the experimental results demonstrated the evaluating capacity of our Reward3D model in assessing the quality of 3D assets as well as its capability in evaluating text-to-3D models, showing that its assessment aligns with human preferences. Consequently, it can serve as a substitute for human evaluation. Both qualitative evaluations and illustrative examples of representative cases have been presented in this part. Please refer to the supplementary for ablation studies.

\begin{figure}[tb]
  \centering
  \includegraphics[width=1\linewidth]{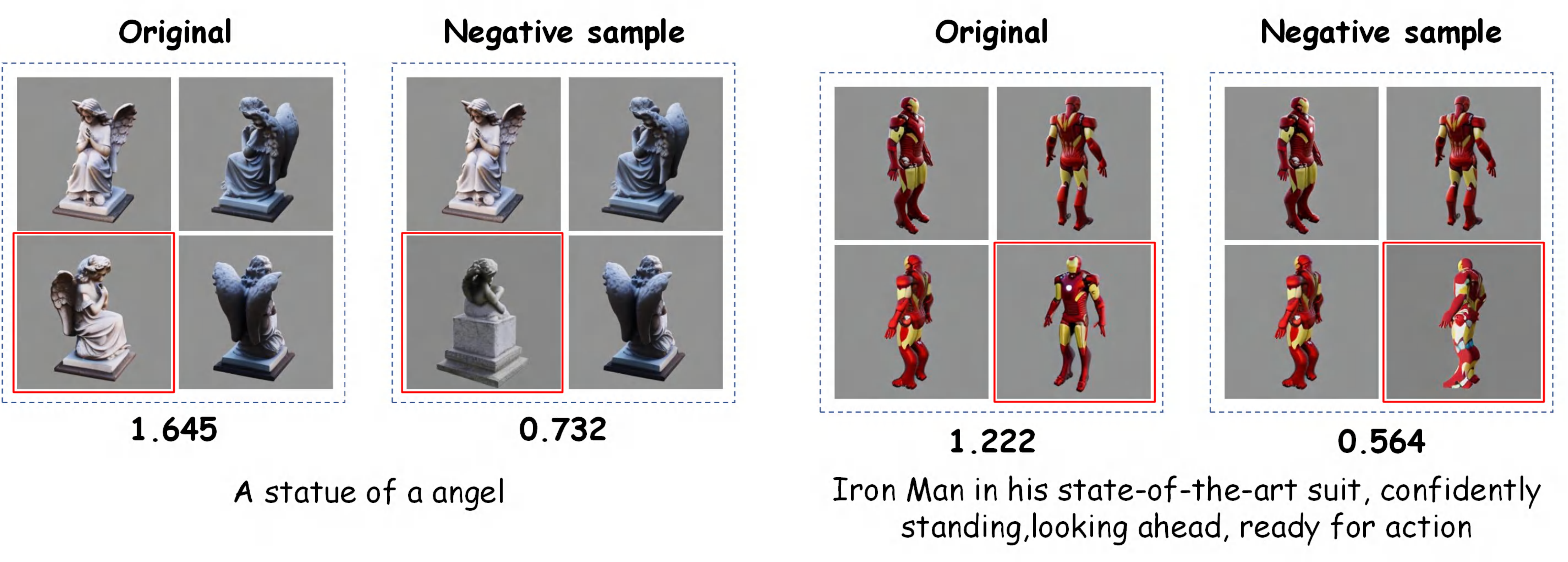}
  \caption{The utilization of Reward3D in scoring both positive examples and negative examples (\textit{left}: inconsistency, \textit{right}: multi-face issue) reveals that the model can effectively distinguish negative examples.}
  \label{fig:result4}
\end{figure}

\begin{table*}[!h]
    \centering
    \caption{\textbf{Quantitative comparisons} on 110 prompts generated by GPTEval3D\cite{wu2024gpt4vision}. We compared our DreamReward with DreamFusion\cite{poole2022dreamfusion}, ProlificDreamer\cite{wang2023prolificdreamer}, Latent-NeRF\cite{metzer2022latentnerf}, MVDream\cite{shi2023mvdream}  , and Fantasia3D\cite{chen2023fantasia3d}. We calculate CLIP$\uparrow$ \cite{ramesh2022hierarchical}, ImageReward$\uparrow$ \cite{xu2023imagereward}, GPTEval3D $\uparrow$ \cite{wu2024gpt4vision}and Reward3D$\uparrow$. Meanwhile, to demonstrate the alignment of our trained Reward3D model with human preferences, we annotated all generated results by researchers. The ranking results in the table below indicate a high alignment between our Reward3D model and the annotated results.}
    \resizebox{\textwidth}{!}{%
\begin{tabular}{ c|c c|c c|c c|c c|c c}
\hline \multirow{3}{*}{ Dataset \& Model } & \multicolumn{10}{|c}{ 110 Prompts from GPTEval3D\cite{wu2024gpt4vision} } \\
\hline & \multicolumn{2}{|c}{ Human Eval }& \multicolumn{2}{|c|}{ GPTEval3D } & \multicolumn{2}{|c|}{ Reward3D } & \multicolumn{2}{|c}{ ImageReward } & \multicolumn{2}{|c}{ CLIP }\\
\hline & Rank & Win & Rank & Score & Rank & Score& Rank & Score& Rank & Score \\
\hline DreamFusion\cite{poole2022dreamfusion}      & 6 & 97  & 6 & 1000 & 6 & -1.597& 5 & -1.489& 5 & 0.224 \\
\hline Fantasia3D\cite{chen2023fantasia3d}       & 5 & 167 & 5 & 1006 & 5 & -1.582& 6 & -1.521& 6 & 0.222 \\
\hline ProlificDreamer\cite{wang2023prolificdreamer}  & 4 & 246 & 4 & 1152 & 4 & -0.195& 4 & -0.639& 3 & 0.252 \\
\hline Latent-NeRF\cite{metzer2022latentnerf}      & 3 & 287 & 3 & 1173 & 3 & -0.012& 2 & -0.350& 2 & 0.257 \\
\hline MVDream\cite{shi2023mvdream}         & 2 & 375 & 2 & 1224 & 2 & 0.246 & 3 & -0.541& 4 & 0.243 \\
\hline Spearman $\rho$ to Human Eval. & \multicolumn{2}{|c}{ - }& \multicolumn{2}{|c|}{ \textbf{1.00} } & \multicolumn{2}{|c|}{ \textbf{1.00} } & \multicolumn{2}{|c}{ 0.80 } & \multicolumn{2}{|c}{ 0.60 }\\
\hline \textbf{DreamReward(Ours)}     & \textbf{1} & \textbf{478} & \textbf{1} & \textbf{1480} & \textbf{1} & \textbf{2.594} & \textbf{1} & \textbf{1.833}& \textbf{1} & \textbf{0.274} \\
\hline
\end{tabular}
    }
    \label{tab:quantitative}
\end{table*}

\begin{figure}[!h]
  \centering
  \includegraphics[width=1\linewidth]{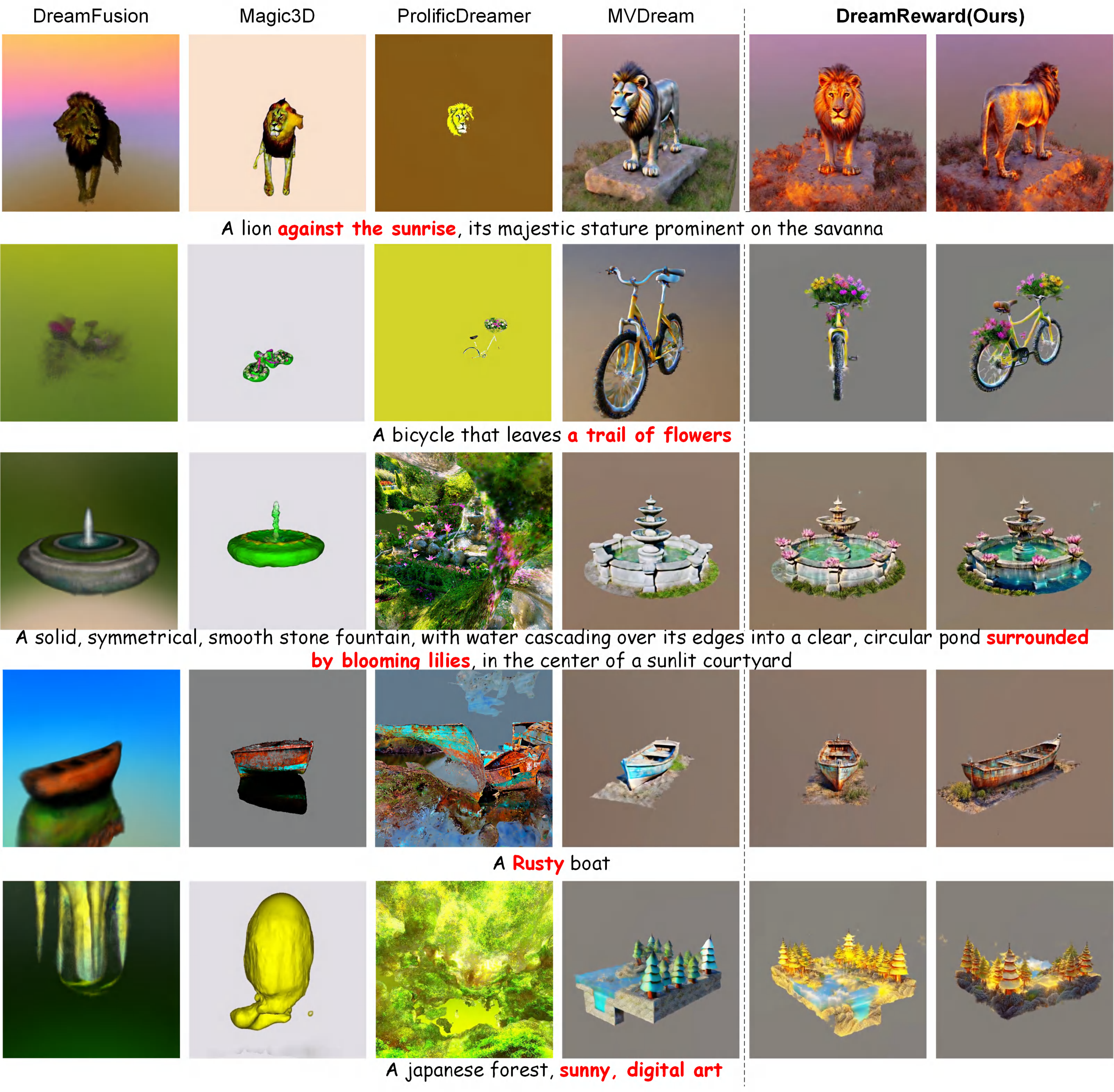}
  \caption{Comparison with four baselines. The results indicate that existing 3D generation models do not align well with human preferences (as highlighted in red). Conversely, our DreamReward results conform more closely to human preferences.}
  \label{fig:result1}
\end{figure}

\begin{figure}[!h]
  \centering
  \includegraphics[width=1\linewidth]{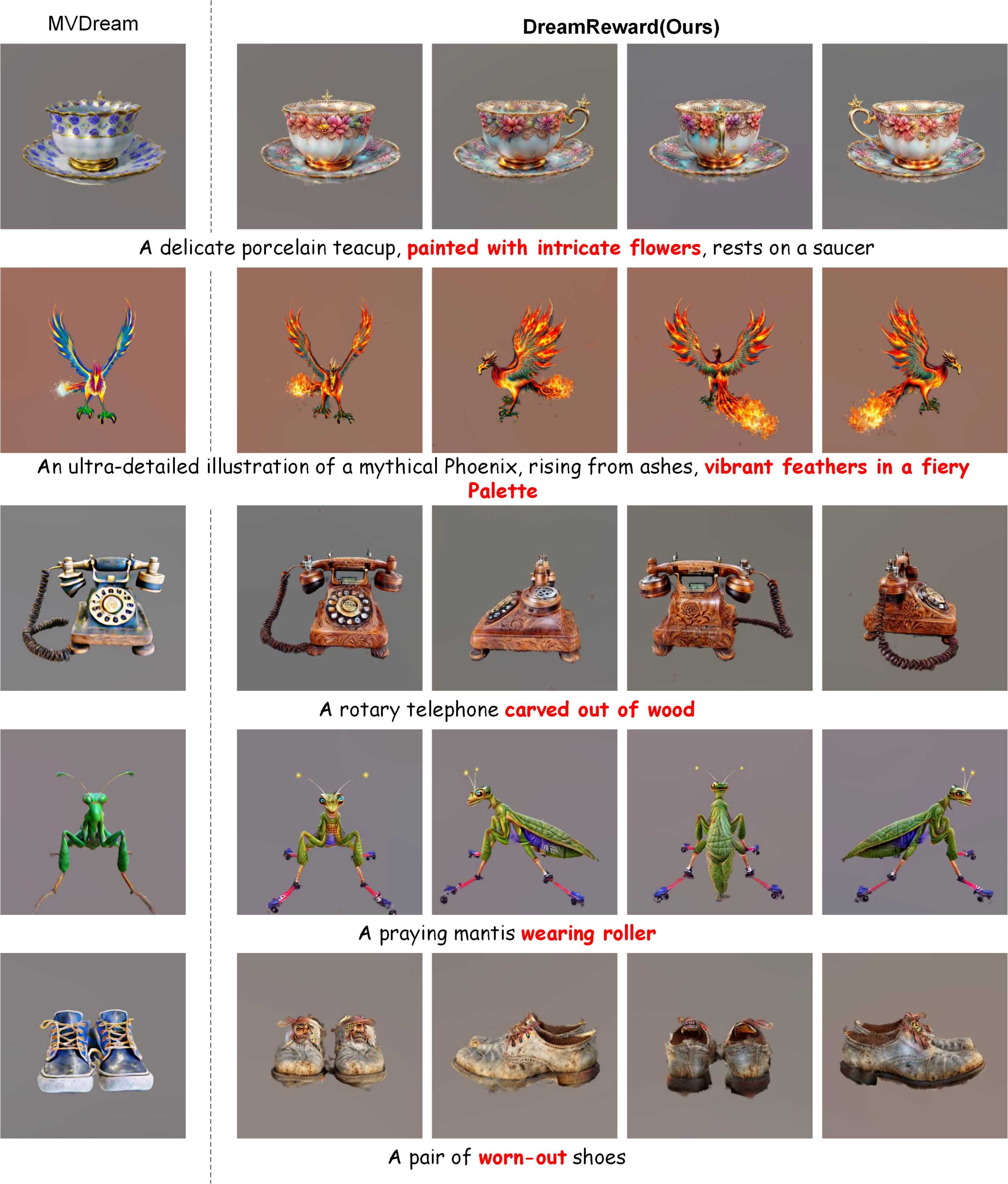}
  \caption{More generated results using our DreamReward. Our work can generate 3D assets of higher alignment, while maintaining consistency across multiple perspectives.}
  \label{fig:result2}
\end{figure}

\begin{figure}[!h]
  \centering
  \includegraphics[width=1\linewidth]{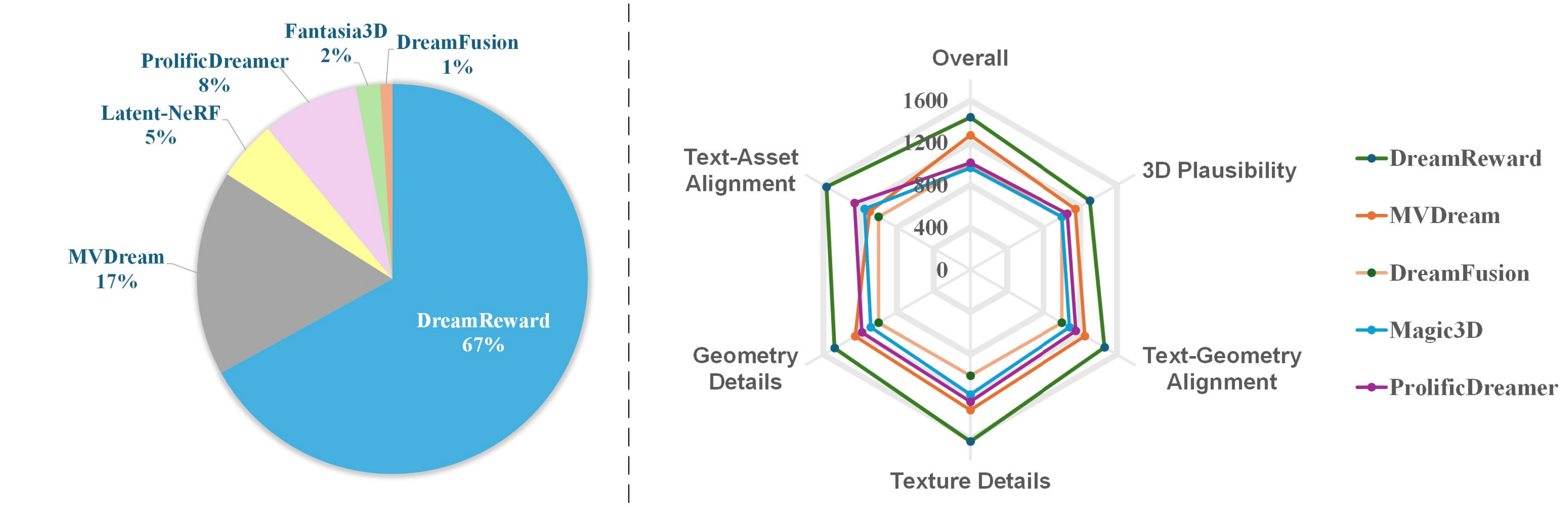}
  \caption{\textbf{Left}: User study of the rate from volunteers’ preference for each method in the inset pie chart, \textbf{Right}: Holistic evaluation using GPTEval3D. The Radar charts report the Elo rating for each of the 6 criteria. The results indicate that our results consistently rank first across all metrics.}
  \label{fig:more comparison}
\end{figure}

\subsection{Experiment Setup}
Our experiments can be divided into two parts: (1) Comparative experiments on DreamReward Sec.~\ref{sec:6.2}, and (2) Comparative experiments on Reward3D Sec.~\ref{sec:6.3}.

In Sec.~\ref{sec:6.2}, we compare our proposed DreamReward with five baseline 3D models: DreamFusion\cite{poole2022dreamfusion}, ProlificDreamer\cite{wang2023prolificdreamer}, Latent-NeRF\cite{metzer2022latentnerf}, MVDream\cite{shi2023mvdream}  , and Fantasia3D\cite{chen2023fantasia3d}. To align with the test results provided in GPTEval3D\cite{wu2024gpt4vision}, we utilize the official implementations of each method when available. Alternatively, we turn to threestudio's\cite{threestudio2023} implementation. The comparative results of the visualization are presented.
Simultaneously, to obtain convincing results, 
four evaluation metrics are used here: CLIP\cite{ramesh2022hierarchical}, GPTEval3D\cite{wu2024gpt4vision}, ImageReward\cite{xu2023imagereward}, and our Reward3D. For the test dataset, we use 110 prompts generated by GPTEval3D, consisting of prompts with varying levels of creativity and complexity. Additionally, we conduct a user study to further demonstrate the alignment of our method with human preferences.

In Sec.~\ref{sec:6.3}, we conduct detailed comparative experiments and user studies on the aforementioned prompt set and 3D baselines. In the course of assessing these models, a large amount of results indicate that our Reward3D better aligns with human preferences compared to existing methods.

\subsection{DreamFL}
\label{sec:6.2}
\subsubsection{Qualitative Comparison.}
Figure~\ref{fig:result1} and Figure~\ref{fig:result2} show the 3D assets generated by four baselines for multiple prompts, allowing for intuitive visual comparison. We observe that the generated results using DreamFusion \cite{poole2022dreamfusion}, ProlificDreamer \cite{wang2023prolificdreamer} and Magic3D \cite{lin2023magic3d} deviate from the text content and also suffer from multi-face problems. While MVDream\cite{shi2023mvdream} can generate high-quality 3D assets with visual consistency, its results still deviate from the given prompt content (as indicated by the red-highlighted text). In comparison, our model can generate 3D assets that align closely with the given prompt while maintaining visual consistency and meeting human aesthetic preferences.
\begin{table*}
    \centering
    \caption{Quantitative comparisons on the alignment, quality and multi-view consistency score in a user study, rated on
a scale of 1-6, with higher scores indicating better performance.}
\begin{tabular}{ c cc c c}
\hline & Alignment$\uparrow$ & Quality$\uparrow$ & Consistency$\uparrow$ & Average$\uparrow$\\
\hline DreamFusion\cite{poole2022dreamfusion}       & 2.65   & 1.95  & 2.95 & 2.52 \\
\hline Fantasia3D\cite{chen2023fantasia3d}       & 2.88  & 3.23  & 2.50 & 2.87 \\
\hline ProlificDreamer\cite{wang2023prolificdreamer}  & 3.90  & 3.78  & 3.13 & 3.60\\
\hline Latent-NeRF\cite{metzer2022latentnerf}      & 3.45  & 3.10  & 3.18 & 3.24 \\
\hline MVDream\cite{shi2023mvdream}         & 3.88 & 4.40  & \textbf{5.38}  & 4.55 \\
\hline \textbf{DreamReward(Ours)}     & \textbf{4.88}  & \textbf{5.03} & 5.30  & \textbf{5.07}\\
\hline
\end{tabular}
    \label{tab:User study}
\end{table*}

\subsubsection{Quantitative Comparison.}
In Table~\ref{tab:quantitative}, we compare our DreamReward with five baselines. It indicates that our results consistently outperform other baseline models across multiple evaluation criteria. In Figure~\ref{fig:more comparison}, we demonstrate that using GPTEval3D\cite{wu2024gpt4vision}, our results consistently outperform other baseline models across all 6 criteria: Text-asset alignment, 3D plausibility, Texture details, Geometry details, Texture-geometry coherency, and overall. For the user study, we extract 120 images from the 60-degree rotating videos rendered by threestudio\cite{threestudio2023}. Each annotator randomly receives 5 sets of multi-view images generated by random methods and is asked to rate them on three aspects: multi-view consistency, consistency between text and model, and personal preference. In addition, they should also choose the favorite one. Finally, we collect the results from 30 participants on 20 text prompts, as shown in Table~\ref{tab:User study} and Figure~\ref{fig:more comparison}. In Table~\ref{tab:User study} we observe that most users consider our results with the highest alignment, generation quality and second consistency. In Figure~\ref{fig:more comparison} we observe that our DreamReward is preferable (65\%) by the raters on average.
\subsubsection{Ablation Study.}
\begin{figure}[tb]
  \centering
  \includegraphics[height=7.5cm]{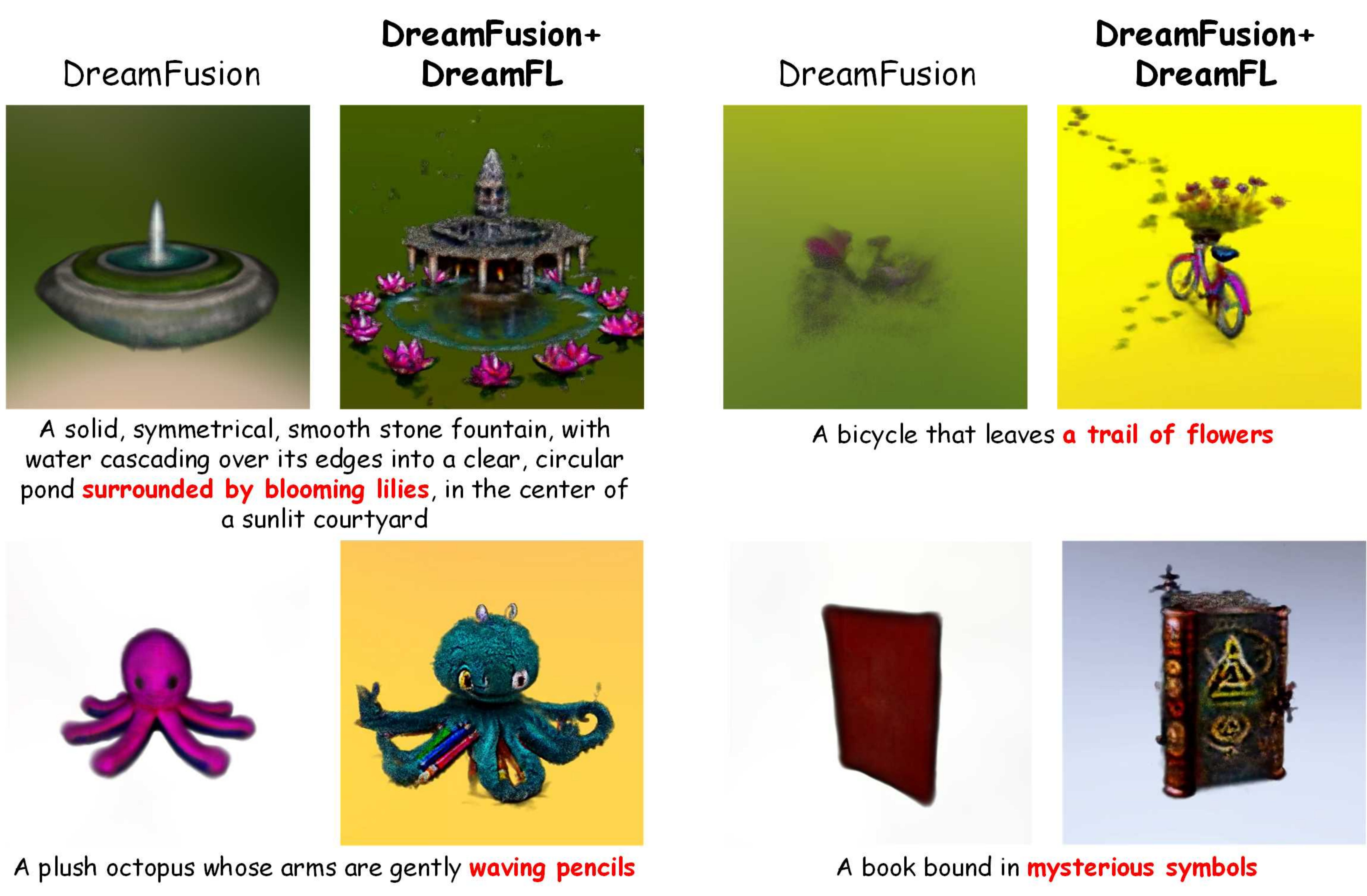}
  \caption{\textbf{Ablation study}. we change the backbone of our DreamFL to DreamFusion\cite{poole2022dreamfusion} and select \emph{stabilityai/stable-diffsuion-2-1-base} \cite{rombach2022highresolution} for 2D diffusion model. We can observe that our method still generates 3D content with higher quality and text alignment.}
  \label{fig:result3}
\end{figure}

We conduct an ablation study based on the backbone. To verify the strength of our method, we implement our DreamFL algorithm on the basis of the DreamFusion architecture. For a fair comparison, we chose the same 2D diffusion model \emph{stabilityai/stable-diffsuion-2-1-base}\cite{rombach2022highresolution}. The results of Figure~\ref{fig:result3} indicate that even for backbones with average performance, incorporating our DreamFL algorithm can achieve better generation quality and text alignment

\subsection{Evaluation of Reward3D}
\label{sec:6.3}
\subsubsection{Quantitative Comparison.} We evaluate the capability of our Reward3D model to determine whether it possesses judgment abilities that align with human aesthetics. This serves to substantiate the feasibility of employing the Reward3D model instead of human evaluation for assessing the quality of 3D assets. The results presented in Table~\ref{tab:quantitative} indicate that the Reward3D model's assessments of 3D assets generated by various models are consistent with human aesthetics, whereas ImageReward \cite{xu2023imagereward} and CLIP \cite{radford2021learning} are not. It is also noteworthy that the results in Table~\ref{tab:quantitative} reveal that GPTEval3D's~\cite{wu2024gpt4vision}(a non-feature model using GPT-4V ~\cite{achiam2023gpt-4v}) evaluating competency is also nearly identical to human assessments. However, in comparison to GPT-4V, our Reward3D model is exceptionally lightweight and offers far faster inference speeds, while still retaining strong evaluation abilities for 3D assets. This suggests a significant advantage of utilizing our Reward3D model for evaluating text-to-3D models as well as for the DreamFL method.
\subsubsection{Representative Cases.}
We further present more visualization results with representative cases. In Figure~\ref{fig:result4}, We select two 3D results and replace one of the viewpoint images with incorrect content, thus creating two negative examples (left: inconsistency, right: multi-face issue). To ease identification, we mark the altered viewpoint images with a red border. We observe that our Reward3D gives lower scores to negative examples. 
These results show that Reward3D accurately assesses 3D content based on human preference and strong 3D awareness. 

\section{Conclusion}
\label{sec:Conclusion}
In this paper, we propose a novel text-to-3D framework called DreamReward for human preference alignment. To the best of our knowledge, we are among the first to utilize RLHF for 3D generation. Specifically, we first construct a new 3D dataset annotated with human preferences consisting of 2530 prompts. Then we train a 3D evaluation model Reward3D on this dataset to better align with human preferences. Powered by the Reward3D model, we further introduce our DreamFL algorithm, which achieves high-fidelity, multi-view consistent, and faithfully human preference-aligned text-to-3D generation. Extensive quantitative and qualitative experiments verify that our DreamReward framework can generate 3D assets with strong human preference alignment.

\noindent \textbf{Limitations and future work.}
Although our proposed DreamReward can generate high-quality and preference-aligned 3D results, there remain requirements for further improvement in the diversity limited by the size of annotated 3D dataset. In our future
work, we will continue to optimize our Reward3D on larger datasets and attempt to incorporate more cameras and orientation information into the Reward3D architecture for better performance.
\bibliographystyle{splncs04}
\bibliography{egbib}
\clearpage
\appendix

\thispagestyle{empty}
\section{The derivation of DreamReward}
In this section, we provide a more detailed derivation of \textbf{DreamReward}. First, let me begin by revisiting the core assumption of this paper:
\begin{equation}
\label{eq1-supp}
\nabla_\theta \mathcal{L}_{\mathrm{Reward}}(\theta) \approx \mathbb{E}_{t, \boldsymbol{\epsilon}, c}\left[\omega(t)\left(\boldsymbol{\epsilon}^r_{\phi}\left(\boldsymbol{x}_t, t, y\right)-\boldsymbol{\epsilon}\right) \frac{\partial \boldsymbol{g}(\theta, c)}{\partial \theta}\right]
\end{equation}
Where $\boldsymbol{\epsilon}^r_{\phi}\left(\boldsymbol{x}_t, t, y\right)$ denotes an ideal noise prediction network aligned with human preference. We approximate it using Reward3D and $\boldsymbol{\epsilon}_{\phi}\left(\boldsymbol{x}_t, t, y\right)$, as follows:
\begin{equation}
    \label{eq2-supp}
    \boldsymbol{\epsilon}^r_{\phi}\left(\boldsymbol{x}_t, t, y\right)=\boldsymbol{\epsilon}_{\phi}\left(\boldsymbol{x}_t, t, y\right)-\delta \epsilon
\end{equation}
The difference $\delta \epsilon$ between the $\epsilon^r_{\phi}(x_t,t,y)$  and $\epsilon_{\phi}(x_t,t,y)$  can be calculated by:
\begin{equation}
\label{eq3-supp}
\delta \epsilon=-\frac{\partial{r(y,\hat{\boldsymbol{x}}_t,c)}}{\partial{\hat{\boldsymbol{x}}_t}}\frac{\partial{\hat{\boldsymbol{x}}_t}}{\partial{\epsilon_{\phi}}}=\lambda \frac{\partial{r(y,\hat{\boldsymbol{x}}_t,c)}}{\partial{\hat{\boldsymbol{x}}_t}}\frac{\partial{\hat{\boldsymbol{x}}_t}}{\partial{\boldsymbol{x}}}=\lambda \frac{\partial{r(y,\hat{\boldsymbol{x}}_t,c)}}{\partial{\boldsymbol{x}}}
\end{equation}
Where $\hat{\boldsymbol{x}}_t$ represents the predicted $\boldsymbol{x}_0$ using $x_t$. In Eq.~\ref{eq3-supp}, $\lambda$ encompasses all constant terms to ensure the equation holds. In practical implementation, we replace $\lambda$ with $\lambda_r$, defined as follows:
\begin{equation}
\label{eq4-supp}
\lambda_r = \mathrm{t}\left(\mathcal{L}_{\mathrm{SDS}}, r\left(y, \hat{\boldsymbol{x}}_t, c\right)\right)\times \mu
\end{equation}
Where $\mu$ denotes a constant hyperparameter, and $\mathrm{t}$ represents the weighting function. Therefore, Eq.~\ref{eq1-supp} can be reorganized as:
\begin{equation}
\label{eq5-supp}
\nabla_\theta \mathcal{L}_{\mathrm{Reward}}(\theta) \approx \mathbb{E}_{t, \boldsymbol{\epsilon}, c}[\omega(t)( \underbrace{\boldsymbol{\epsilon}_{\phi}\left(\boldsymbol{x}_t, t, y\right)-\lambda_r \frac{\partial{r(y,\hat{\boldsymbol{x}}_t,c)}}{\partial{g(\theta,c)}}}_{\boldsymbol{\epsilon}_{\phi}^r\left(\boldsymbol{x}_t, t, y\right)}-\boldsymbol{\epsilon}) \frac{\partial \boldsymbol{g}(\theta, c)}{\partial \theta}].
\end{equation}
SDS can be reorganized as:
\begin{equation}
\label{eq6-supp}
\mathcal{L}_{\mathrm{Reward}}(\theta) \approx \mathcal{L}_{\mathrm{SDS}}(\theta)-\lambda_{r} r(y,\hat{\boldsymbol{x}_t},c).
\end{equation}
\section{Additional Implementation Details}
\subsection{Pseudo-code for DreamFL}
A more detailed pseudo-code for DreamFL is presented in~\cref{alg:DreamReward}.
\begin{algorithm}
\caption{Pseudo-code for DreamFL}
\label{alg:DreamReward}

\textbf{Input: }{ Large pretrained text-to-image diffusion model 
$\mathbf{\epsilon}_{\phi}$. Reward3D model $r$. Learning rate $\eta_{1}$ for 3D structures parameters. A prompt $y$. The weight of reward loss $\lambda_r$}. Evaluating threshold of noising time $t_{\text{threshold}}$.
\begin{algorithmic}[1]
\STATE {\bfseries initialize}  A 3D structure presenting with NeRF $\theta$.
\WHILE{not converged}
\STATE Randomly sample a camera pose $c$, 2D noise $\vepsilon\sim\mathcal{N}(\vzero,I)$, and timestep $t\sim \mathrm{Uniform}(\{1, \dotsc, T\})$.
\STATE Render at pose $c$ to get a multiview image $\vx_0=\vg(\theta,c)\in \mathbb{R}^{4\times H \times W \times C}$.
\IF{$t < t_{threshold}$}
\STATE Add noise $\vepsilon$ to $\vx_0$ and get $\vx_t$.
\STATE Denoise with the predicting noise $\hat\vx_t = \left( \vx_t - \sqrt{1-\bar\alpha_t}\vepsilon_\phi(\vx_t,t,y) \right) / \sqrt{\bar\alpha_t}\mbox{.}$
\STATE $\mathcal{L}_{\text{Reward}}=\mathcal{L}_{\text{SDS}}-\lambda_r r(y,\hat\vx_t,c)\mbox{.} $
\STATE $\delta\epsilon \leftarrow \lambda_r \nabla_{\boldsymbol{g}(\theta, c)}r$
\ELSE
\STATE $\mathcal{L}_{\text{Reward}}=\mathcal{L}_{\text{SDS}}$
\STATE $\delta\epsilon \leftarrow 0.$
\ENDIF
\STATE $\nabla_\theta \mathcal{L}_{\text{Reward}}(\theta) \approx \mathbb{E}_{t, \boldsymbol{\epsilon}, c}\left[\omega(t)\left(\boldsymbol{\epsilon}_{\phi}\left(\boldsymbol{x}_t, t, y\right)-\delta\epsilon-\boldsymbol{\epsilon}\right) \frac{\partial \boldsymbol{g}(\theta, c)}{\partial \theta}\right]$
\STATE $\theta \leftarrow \theta - \eta\nabla_{\theta}\mathcal{L}\mbox{.} $
\ENDWHILE
\STATE {\bfseries return} 
\end{algorithmic}
\end{algorithm}
\subsection{Training Details}
The time consumption for our DreamReward is 40 minutes (which is similar to MVDream~\cite{shi2023mvdream}). It is worth mentioning that our DreamFL can also be applied to traditional RLHF. Specifically, by utilizing the DreamFL algorithm, a pre-trained 3D asset can be fine-tuned in just 2 minutes. The resulting 3D asset will exhibit finer texture details, more appealing colors, and content that better aligns with textual descriptions.
\section{Additional Experiments}
In this section, we provide more visual examples to demonstrate that our DreamReward is more aligned with human preferences.
\subsection{More Visual Results}
Figure~\ref{fig:supp1}, \ref{fig:supp2} presents additional comparison results with MVDream~\cite{shi2023mvdream}. We adhere to the definition used in GPTEval3D~\cite{wu2024gpt4vision}, dividing prompts into the subject part and the properties part (e.g., in "a sleeping cat," "cat" is the subject while "sleeping" represents the properties). Based on experiments~\cite{shi2023mvdream}, it is observed that MVDream~\cite{shi2023mvdream} can effectively generate the subject part but encounters difficulties with certain more complex properties. For instance, in the "A dancing elephant" depicted in Figure~\ref{fig:supp1}, MVDream~\cite{shi2023mvdream} can depict the elephant accurately but struggles to convey its dancing attribute. On the other hand, our DreamReward can effectively overcome this issue, thus generating high-quality 3D assets that align with the properties described.
\subsection{More Qualitative Comparisons}
\begin{figure*}[!t]
  \centering
  \includegraphics[width=0.94\linewidth]{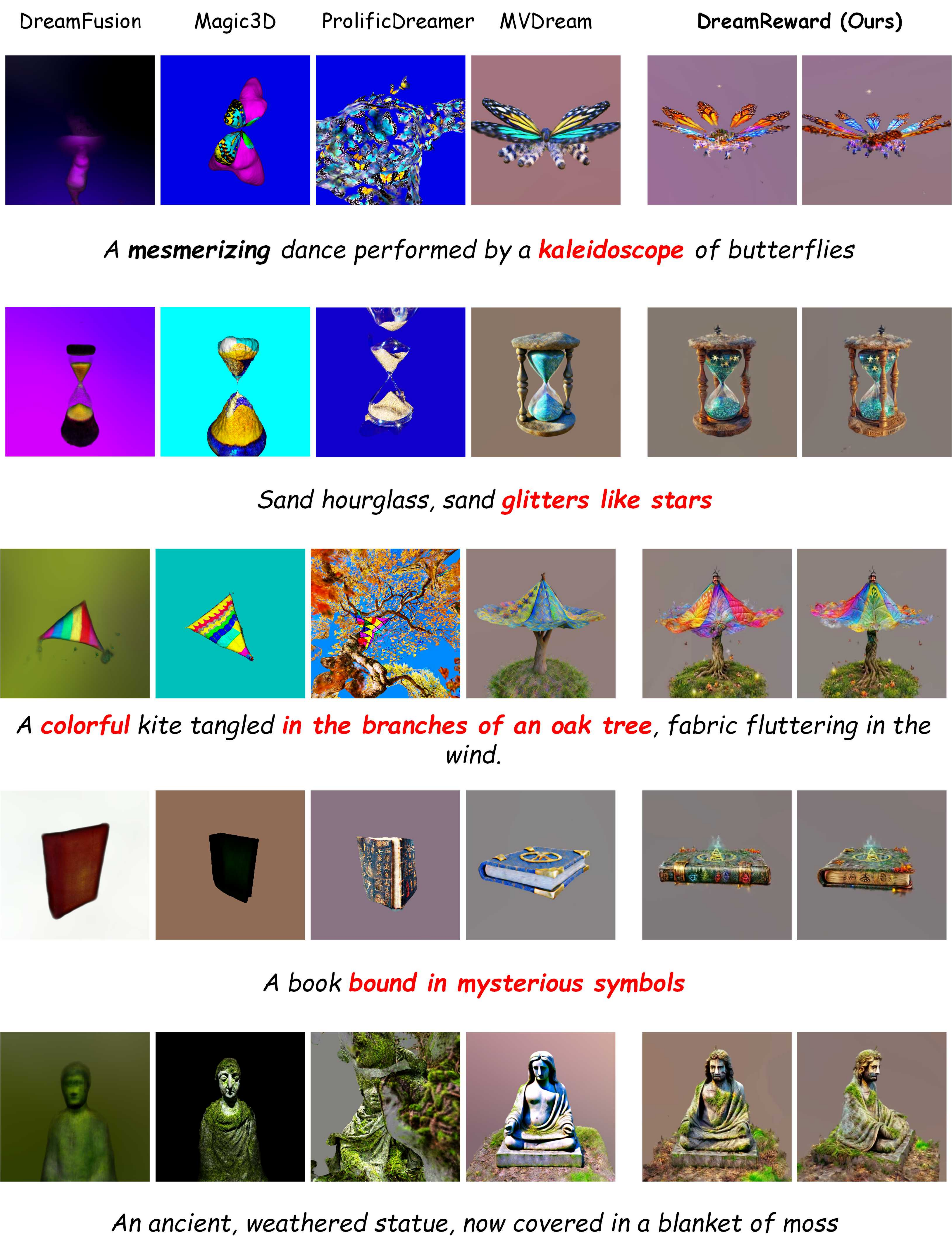}
  \caption{\textbf{More comparisons with four baselines}}
  \label{fig:supp4}
\end{figure*}

Figure~\ref{fig:supp4} presents additional comparison results with DreamFusion~\cite{poole2022dreamfusion}, ProlificDreamer~\cite{wang2023prolificdreamer}, Magic3D~\cite{lin2023magic3d} and MVDream~\cite{shi2023mvdream}. These examples indicate that the properties of the 3D assets generated by baseline models fail to align with the provided prompts. In contrast, our DreamReward not only ensures consistency between text and 3D output but also provides results that are more in line with human aesthetics.
\subsection{More Quantitative Comparison}
When evaluating the 3D assets generated by our DreamReward using GPTEval3D~\cite{wu2024gpt4vision}, we employ the latest official open-source code and run it with default configuration settings to maintain fairness. To ensure comprehensive assessment, we conduct evaluations through two different comparison methods: (1) competing against 13 baseline models and (2) competing solely against MVDream. We conduct 220 and 110 inquiries/competitions respectively. The Radar charts presented in the main text are derived from the results of the first method, while the specific ELo scores extracted from these charts are detailed in Table~\ref{tab:ELO}. Following the completion of evaluation (1), we observe that although we achieve the best results across various metrics, there are instances of structural collapse for certain prompts. Later, we choose a new hyperparameter $\lambda_r$ and uniformly regenerate 110 3D assets using this updated value. We conduct evaluation (2) based on the new 3D assets, with the ELO scores presented in Table~\ref{tab:ELO2}.
\subsection{GPT-4V Comparison Examples}
\begin{figure*}[!t]
  \centering
  \includegraphics[width=0.94\linewidth]{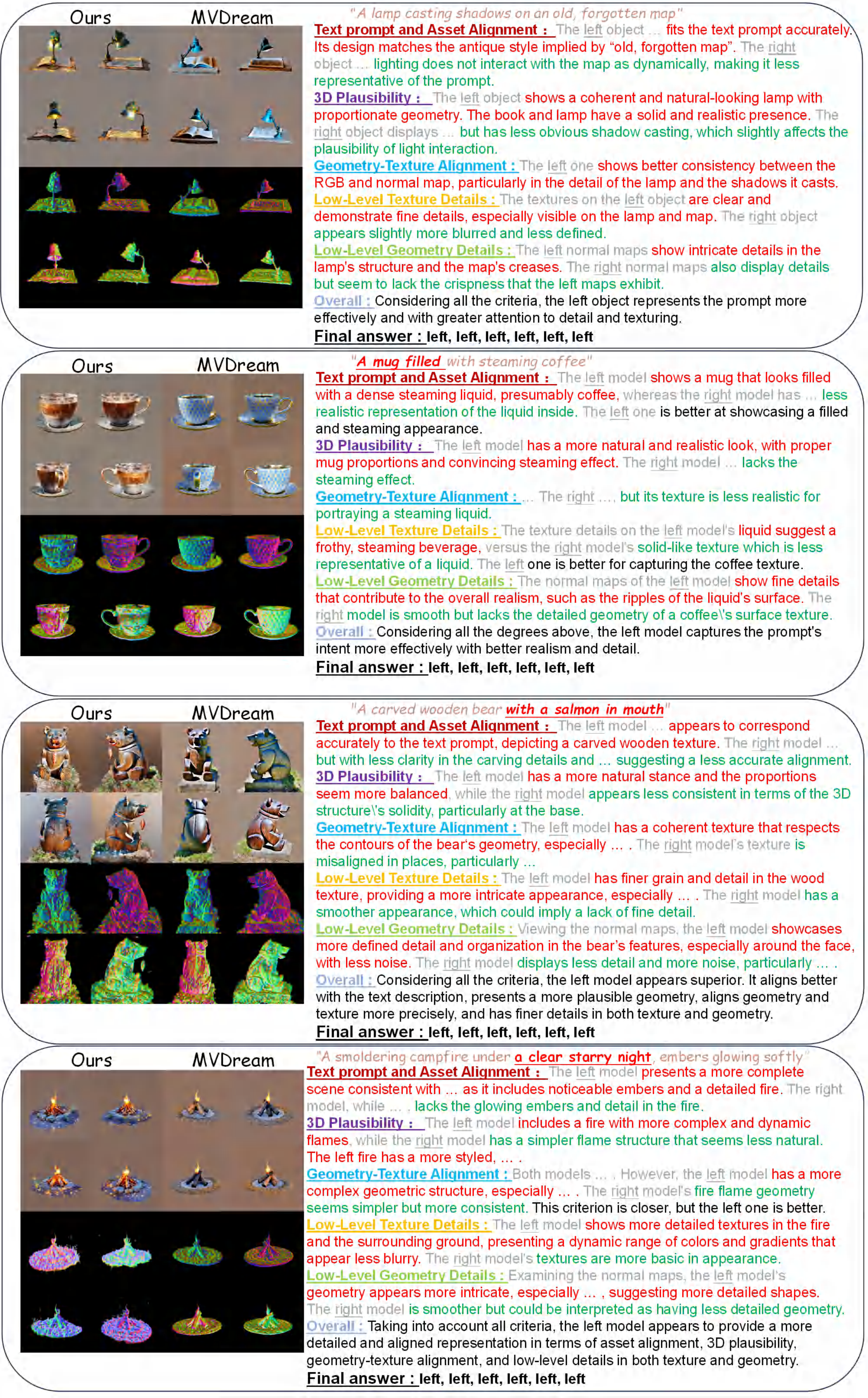}
  \caption{\textbf{Examples of the analysis by GPTEval3D~\cite{wu2024gpt4vision}}}
  \label{fig:gpt1}
\end{figure*}

To better illustrate the higher quality of our generated results, we choose GPTEval3D~\cite{wu2024gpt4vision} for additional pairwise comparison. As shown in Figure~\ref{fig:gpt1}, we present four pairs of comparison results from the tournament~\cite{wu2024gpt4vision}. GPT4V~\cite{achiam2023gpt-4v} scores each pair from six aspects: Text-asset alignment, 3D plausibility, Texture details, Geometry details, Texture-geometry coherence, and overall. 
\begin{table*}
    \centering
    \caption{The first evaluation results using GPTEval3D~\cite{wu2024gpt4vision} correspond to the Radar charts in the main text. We bold the best result and underline the second-best result.}
    \renewcommand{\arraystretch}{1.5}
    \resizebox{\linewidth}{!}{
\begin{tabular}{ c cc c c c c}
\hline & \parbox{2cm}{Text-Asset Alignment} & \parbox{2cm}{3D-Plausibility} & \parbox{2cm}{Text-Geometry Alignment} &\parbox{2cm}{Geometry-Details}& \parbox{1.5cm}{Texture-Details}& Overall\\
\hline DreamFusion~\cite{poole2022dreamfusion}      & 1000   & 1000  & 1000  & 1000 &1000 &1000  \\
\hline Fantasia3D~\cite{chen2023fantasia3d}     & 1068  & 892  & 1006 & 1028&1109&934 \\
\hline ProlificDreamer~\cite{wang2023prolificdreamer} & \underline{1262}  & 1059  & 1152 & 1181&1246&1013\\
\hline Latent-NeRF~\cite{metzer2022latentnerf}      & 1222  & 1145  & 1157 & 1161&1180&1179 \\
\hline MVDream~\cite{shi2023mvdream}         & 1098 & \underline{1148} & \underline{1251}  & \underline{1256}&\underline{1325}& \underline{1271}\\
\hline \textbf{Ours}  & \textbf{1567}  & \textbf{1305} & \textbf{1465}  & \textbf{1479}&\textbf{1622}&\textbf{1442}\\
\hline
\end{tabular}}
    \label{tab:ELO}
\end{table*}
\begin{table*}[!h]
    \centering
    \caption{The second evaluation results using GPTEval3D~\cite{wu2024gpt4vision}. We bold the best result.}
    \renewcommand{\arraystretch}{1.5}
    \resizebox{\linewidth}{!}{
\begin{tabular}{ c cc c c c c}
\hline & \parbox{2cm}{Text-Asset Alignment} & \parbox{2cm}{3D-Plausibility} & \parbox{2cm}{Text-Geometry Alignment} &\parbox{2cm}{Geometry-Details}& \parbox{1.5cm}{Texture-Details}& Overall\\
\hline MVDream~\cite{shi2023mvdream}         & 779 & 815 & 810  & 784&780& 784\\
\hline \textbf{Ours}  & \textbf{1221}  & \textbf{1185} & \textbf{1190}  & \textbf{1216}&\textbf{1220}&\textbf{1216}\\
\hline
\end{tabular}}
    \label{tab:ELO2}
\end{table*}
\subsection{More Ablation Study}
\begin{figure*}[!t]
  \centering
  \includegraphics[width=0.94\linewidth]{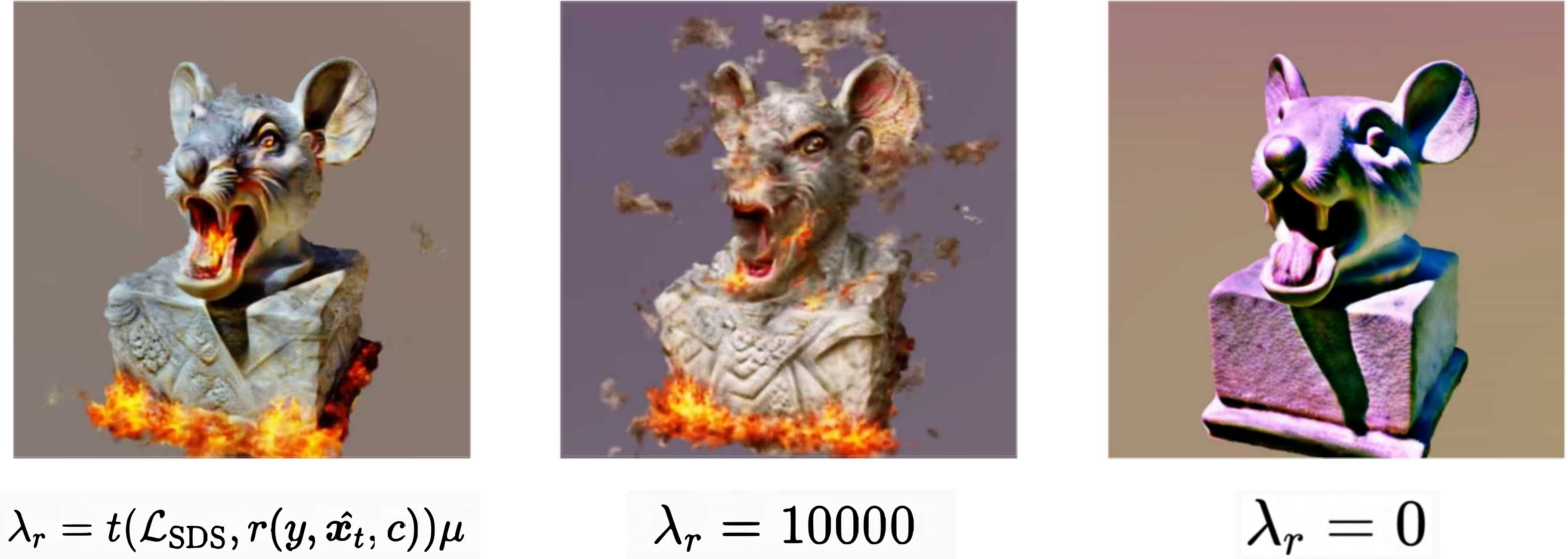}
  \caption{\textbf{More ablation study} about $\lambda_r$, the prompt is: "A marble bust of a mouse with its mouth open, spewing fire."}
  \label{fig:abl1}
\end{figure*}

We conduct ablation study on the weight parameter $\lambda_r$ of the reward. Figure~\ref{fig:abl1} indicate that when $\lambda_r$ is set to 0 (i.e., degenerating into a standard SDS), the generated results only consist of subjects without properties. When $\lambda_r$ is set to a constant 10000, the form of the generated results tends to collapse. However, with the application of our designed weight function t, the generated results maintain both high quality and alignment simultaneously.

\begin{figure*}[!t]
  \centering
  \includegraphics[width=0.93\linewidth]{fig/new1-2.pdf}
  \caption{\textbf{More generated results using our DreamReward}}
  \label{fig:supp1}
\end{figure*}

\begin{figure*}[!t]
  \centering
  \includegraphics[width=0.91\linewidth]{fig/new2-2.pdf}
  \caption{\textbf{More generated results using our DreamReward}}
  \label{fig:supp2}
\end{figure*}


\end{document}